\documentclass[10pt,twocolumn,letterpaper]{article}

\usepackage{btas}
\usepackage{times}
\usepackage{epsfig}
\usepackage{graphicx}
\usepackage{amsmath}
\usepackage{amssymb}
\usepackage{gensymb}
\usepackage{url}

\usepackage{array}
\usepackage{multirow}
\usepackage{textcomp}
\usepackage{array}
\usepackage{xcolor,colortbl}

\btasfinalcopy 

\ifbtasfinal\fi
\begin{document}

\title{Human Iris Recognition in Post-mortem Subjects: Study and Database}

\author{Mateusz Trokielewicz$^{\dag,\ddag}$, Adam Czajka$^{\dag,\ddag}$\\
$^{\dag}$Biometrics Laboratory\\
Research and Academic Computer Network\\
Wawozowa 18, 02-796 Warsaw, Poland\\
$^{\ddag}$Institute of Control and Computation Engineering\\
Warsaw University of Technology\\
Nowowiejska 15/19, 00-665 Warsaw, Poland\\
{\tt\small {mateusz.trokielewicz,adam.czajka}@nask.pl}
\and
Piotr Maciejewicz$^{\star}$\\
$^{\star}$Department of Ophthalmology\\
Medical University of Warsaw\\
Lindleya 4, 02-005 Warsaw, Poland\\
{\tt\small piotr.maciejewicz@wum.edu.pl}
}
\maketitle
\thispagestyle{empty}

\begin{abstract}

This paper presents a unique study of post-mortem human iris recognition and the first known to us database of near-infrared and visible-light iris images of deceased humans collected up to almost 17 days after death. We used four different iris recognition methods to analyze the dynamics of iris quality decay in short-term comparisons (samples collected up to 60 hours after death) and long-term comparisons (for samples acquired up to 407 hours after demise). This study shows that post-mortem iris recognition is possible and occasionally works even 17 days after death. These conclusions contradict a promulgated rumor that iris is unusable shortly after decease. We make this dataset publicly available to let others verify our findings and to research new aspects of this important and unfamiliar topic. We are not aware of any earlier papers offering post-mortem human iris images and such comprehensive analysis employing four different matchers.
\let\thefootnote\relax\footnote{Accepted for publication at the IEEE BTAS2016}
\end{abstract}

\section{Introduction}
The field of biometrics sometimes must address issues that are important, yet also unpleasant. Post-mortem biometric recognition is one such subjects. In particular, iris biometrics in the deceased is a problem that has seen very little research. However, even with little or no published work on the problem, there are still some statements in the literature that assert post-mortem iris recognition as not possible. Revealing the truth is of the utmost importance for at least two reasons, namely forensics and identity protection. The first could prove useful in fast identity verification in victims of accidents, crimes or even in the battlefield during warfare, providing a new tool for forensic examiners, \emph{e.g.}, when fingerprints are not available (victims with severed hands or fingers, individuals with deliberately altered fingerprints). The latter reason relates to social concerns regarding identity theft: \emph{'will someone be able to steal my iris after I die, and use it to gain access to my identity?'} \cite{ScienceFocusPostMortem}. If iris was a preferred biometric modality for the deceased person, this would have significant repercussions regarding safety of the assets protected by it.

This paper presents short-term (for samples acquired up to 60 hours after death) and long-term (65 to 407 hours after decease) analyses of post-mortem iris recognition deterioration dynamics. Results presented in Section \ref{sec:results} show that iris recognition works correctly in selected cases even 17 days after decease, if the body is kept in mortuary conditions. We found that neither the pupil size changes significantly over a period of many days, nor are there any signs of iris tissue 'vanishing'. Also, due to the use of near-infrared (NIR) illumination in the capturing process, the iris is visible even when significant corneal opacities start to appear.

This study had an institutional review board clearance and the ethical principles of the Helsinki Declaration were followed by the authors. The authors also took all required legal efforts to make this unique database publicly available. This enables other researchers to verify all findings presented in this study, as well as find other aspects of post-mortem iris recognition that we might have overlooked. We are not aware of any other dataset of post-mortem iris images, and this study seems to be the most comprehensive in terms of the time lapse after decease, and the number of iris recognition methods applied. Interested researchers are encouraged to follow the instructions given at the webpage\footnote{\url{http://zbum.ia.pw.edu.pl/EN}$\rightarrow$Research$\rightarrow$Databases} to get a copy of this database.

\section{Related work}
A common belief that iris recognition after death would not be possible is repeated in a number of publications, starting in 2001 with Daugman saying that \emph{'soon after death, the pupil dilates considerably, and the cornea becomes cloudy'} \cite{DaugmanPostMortem}. Similar statements are then surfacing in both the scientific literature (\eg, \emph{'the iris (...) decays only a few minutes after death'} \cite{SaeedPostMortem}), as well as in descriptions of commercial systems (\eg, \emph{after death, a person's iris features will vanish along with pupil's dilation'} \cite{IriTechPostMortem}, \emph{'The iris (...) completely relaxes after death and results in a fully dilated pupil with no visible iris at all. A dead person simply does not have a usable iris!'} \cite{IrisGuardPostMortem}). However, none of these claims were supported by experimental evidence. In our previous work on this subject we have shown that statements regarding corneal opacity and excessive pupil dilation are only partially true, and that images obtained 27 hours post-mortem can still be successfully recognized (with accuracy reaching 70\%) \cite{TrokielewiczPostMortemICB2016}. One of this paper's goals is therefore to examine whether such assertions can be considered true or not when images obtained after even longer periods of time are considered.

Saripalle \etal \cite{PostMortemPigs} report on biometric performance of pig irises, with eyes removed from the cadaver. Irises are shown to slowly degrade with time and lose their biometric capabilities after 6 to 8 hours. Sansola \cite{BostonPostMortem} presents an evaluation of human iris recognition using images obtained from the deceased and one commercial iris method, reporting 20\% false non-matches and no false matches when subsequent iris scans are compared to those obtained a moment after the demise. Finally, the CITeR project related to the topic took place, but no scientific papers have been yet released \cite{RossPostMortem}. 

\section{Medical background}

\label{sec:Medical}
To understand the processes occurring in eye tissues after death, one has to become acquainted with underlying biological, anatomical, and biochemical mechanisms of the human body. Changes started at the molecular level (\eg, changes in potassium concentration in the vitreous humor in the posterior chamber) tend to progress sequentially to micro- and macroscopic morphology of the tissues.

The most prominent metamorphoses that may be observed in the eyes after death, and possibly the most troubling for iris recognition, are the changes to the corneal tissue. Cornea, an avascular convex refractile, overlying the anterior chamber of the eye, is fully transparent in its normal condition. This state is maintained by a controlled hydration with the tear film, produced by the lacrimal glands and distributed by eyelid blinking. As the secretion stops, \emph{anoxia} (lack of oxygenation), dehydration and acidosis lead to slow, yet progressing autolysis of the cells. This results in opacification increasing with time. Another effect associated with these mechanisms is the wrinkling of the cornea, manifesting itself with difficulties to obtain a good visibility of the underlying iris tissue. The progress of these effects is heavily influenced by multiple factors: closure of the eyelids, environment humidity, temperature, and air movement. It is also dependent on the age and general medical condition of the deceased person. Due to decrease in the pressure of the vitreous humor normally present in the eyeball, a certain effect of eyeball \emph{collapse} can be also observed, as the eye slowly sinks into the eye cavity and loses its firmness. Despite what is commonly stated in the biometric literature, no severe changes to the iris tissue itself should be observed immediately after death, neither should there be an evident 'relaxation' of the pupil sphincter and dilator muscles. After demise, pupils are fixed in the so called 'cadaveric position', usually mid-dilated. Their shape and size often depend on the previous medical history of the subject, including treatment with certain drugs. Moreover, the iris of a deceased person is expected not to react to light, but can, for a few hours, react to local chemical stimulation.

\section{Database}
\label{sec:DataCollection}

\subsection{Acquisition methodology and timeline}

A new database of iris images representing eyes of recently deceased persons was created for the purpose of this study. Each eye was photographed using two different sensors: a professional, handheld NIR iris recognition camera (IriShield M2120U), and a consumer color camera (Olympus TG-3). Color images, collected together with NIR images in the same timeframe, could help assess the changes occurring in the eye after death and their possible repercussions for iris recognition. The temperature in the hospital mortuary where data collection was performed was approx. 6\degree~Celsius (42.8\degree~Fahrenheit).

\begin{figure}[!htb]
\centering
\includegraphics[width=0.47\textwidth]{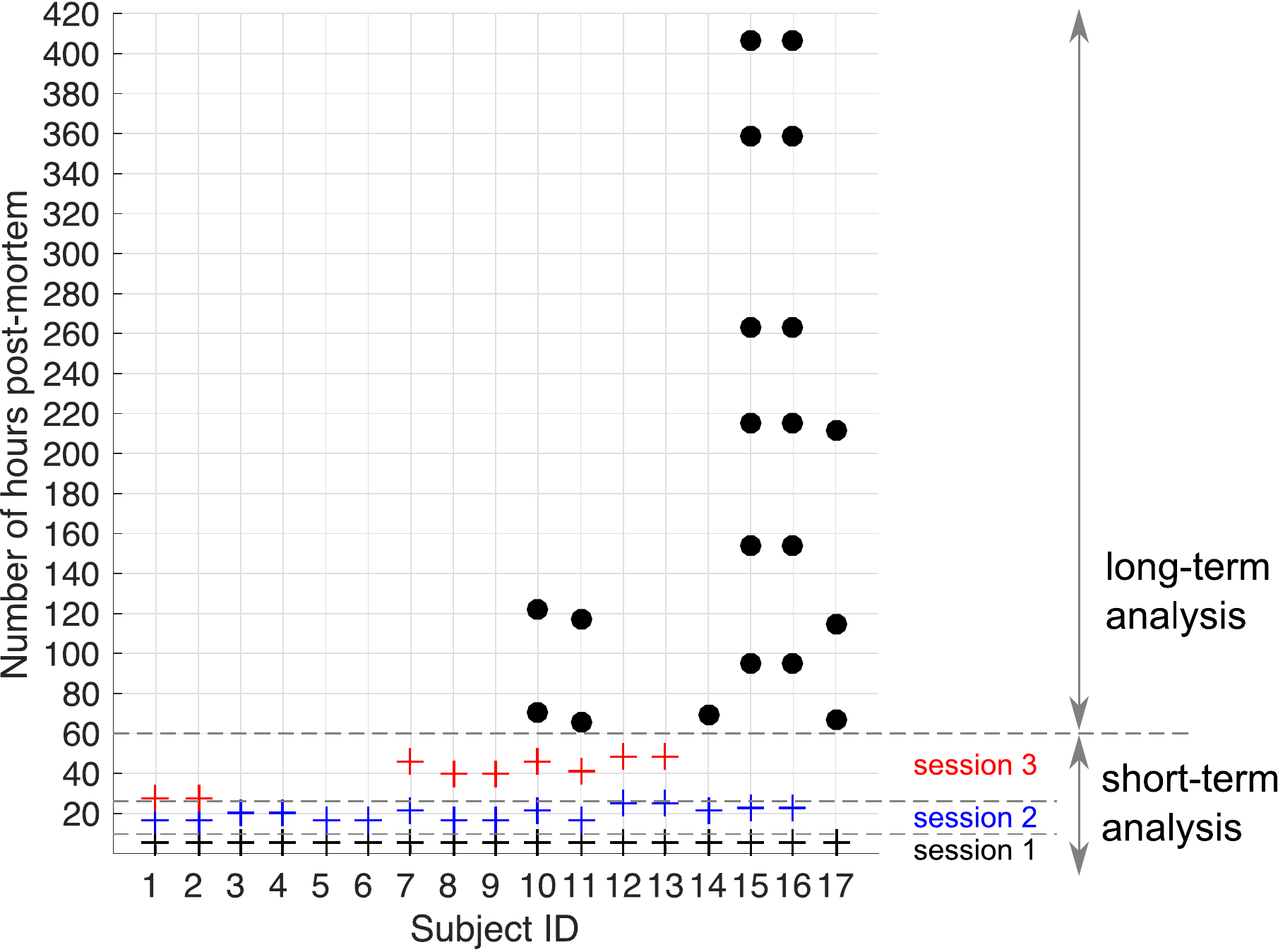}
\caption{Distribution of acquisition moments for all 17 subjects. Each black, blue and red cross represents a single acquisition session included into a short-term analysis. Black dots represent sparse acquisition sessions, organized more than 60 hours after death, and included into a long-term analysis.}
\label{fig:sessions}
\end{figure}

Depending on the tissue availability, images were collected in 2 to 8 acquisition sessions. Single-session images were acquired in separate presentations, as recommended by ISO/IEC 19795-2, \emph{i.e.}, the camera was moved away from the subject and positioned again for the next acquisition. The first session was always conducted approximately 5--7 hours after death. The second-session images were taken 16--27 hours after death. The third session took place 27.5--60 hours after demise. Remaining 5 sessions were more sparsely and unevenly distributed in time and across subjects, Fig. \ref{fig:sessions}. Thus, the analysis presented further in this study was done for short-term data, encompassing samples collected up to 60 hours after death, and long-term data, which gathers all the remaining samples.

\subsection{Database statistics}

The full dataset collected for this study consists of 480 NIR-illuminated images accompanied by 850 color photographs. Images come from 34 distinct irises (17 subjects). Age of the deceased ranged from 37 to 75 years old. 4 subjects were female and 13 were male. Causes of death included circulatory failure (9 subjects), traffic accident (4), suicide by hanging (2), murder (1), and poisoning (1). Detailed description for each subject can be found in the metadata accompanying the released database.

\subsection{Visual inspection}

\label{sec:visualInspection}
\begin{figure}[!htb]
\centering
\includegraphics[width=0.47\textwidth]{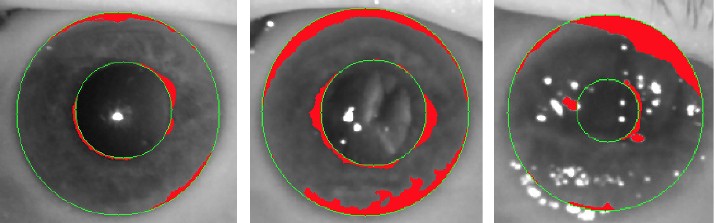}
\caption{Close-ups of iris area with OSIRIS segmentation for samples of subject \#15 acquired in session 1 (left), session 3 (middle) and session 8 (right) with visible wrinkles in the cornea and iris.}
\label{fig:coldSamples}
\end{figure}

When performing a visual inspection of the samples, several interesting observations can be made. First, NIR and visible light illumination reveal different appearance of the iris, and NIR illumination seems to better penetrate corneal occlusions that obstruct the iris pattern, consistently with reports \cite{Aslam,TrokielewiczBTAS2015}, in which NIR illumination is shown to offer good visibility of iris tissue in spite of some degree of opacification. Second, in selected cases a post-mortem loss of intra-ocular pressure is present, which results in a partial collapse of the eyeball leading to significant changes in the appearance of the iris. This can be compensated for (to a certain extent) by applying physical pressure to the eyeball. Third, with increasing time interval since the subject's demise, a wrinkling of the cornea begins, caused by drying of its surface. Such degradation can be expected to affect the recognition accuracy, as the iris pattern behind the wrinkled cornea appears to change significantly, Fig. \ref{fig:coldSamples}. Finally, the pupil shape and size also change as time since death elapses. Those differences, however, are rather small or -- in selected cases -- not present at all.

\section{Iris recognition tools}
This study employs four different, well known iris recognition methods: IriCore (IriTech Inc., \cite{IriCore}), VeriEye (Neurotechnology, \cite{VeriEye}), MIRLIN (formerly SmartSensors, now Fotonation, \cite{MIRLIN}) and OSIRIS (an open source method developed in the BioSecure European project, \cite{OSIRIS}). IriCore and VeriEye implement unpublished iris recognition methods. IriCore and the iris sensor used in this study come from the same manufacturer, thus the IriCore results should be taken with a greater confidence. It returns a dissimilarity between samples, where 0.0 denotes perfect match and 2.0 denotes perfect non-match. VeriEye gives similarity between irises: the higher the score, the better the match, and 0.0 denotes perfect non-match. MIRLIN calculates a DCT-based binary code for the overlapping angular iris image patches and calculates fractional Hamming distance between codes. Values close to zero are expected for same-eye images, and values around 0.5 for different eyes. OSIRIS implements Daugman's idea based on 2D Gabor wavelets, and -- like in MIRLIN -- fractional HD is used to calculate the match. Apart from the OSIRIS method, which does not implement any presentation attack detection (PAD), all of the SDKs and the IriShield camera used in this research were closed products with no detailed information about the PAD possibly realized by the software. If there was any PAD active in the camera, it was not based on pupil dynamics, as we were able to acquire images for eyes with static pupils.

\section{Results}
\label{sec:results}
\subsection{Short-term analysis}

\begin{figure*}[!htb]
	\centering
	\includegraphics[width=0.46\textwidth]{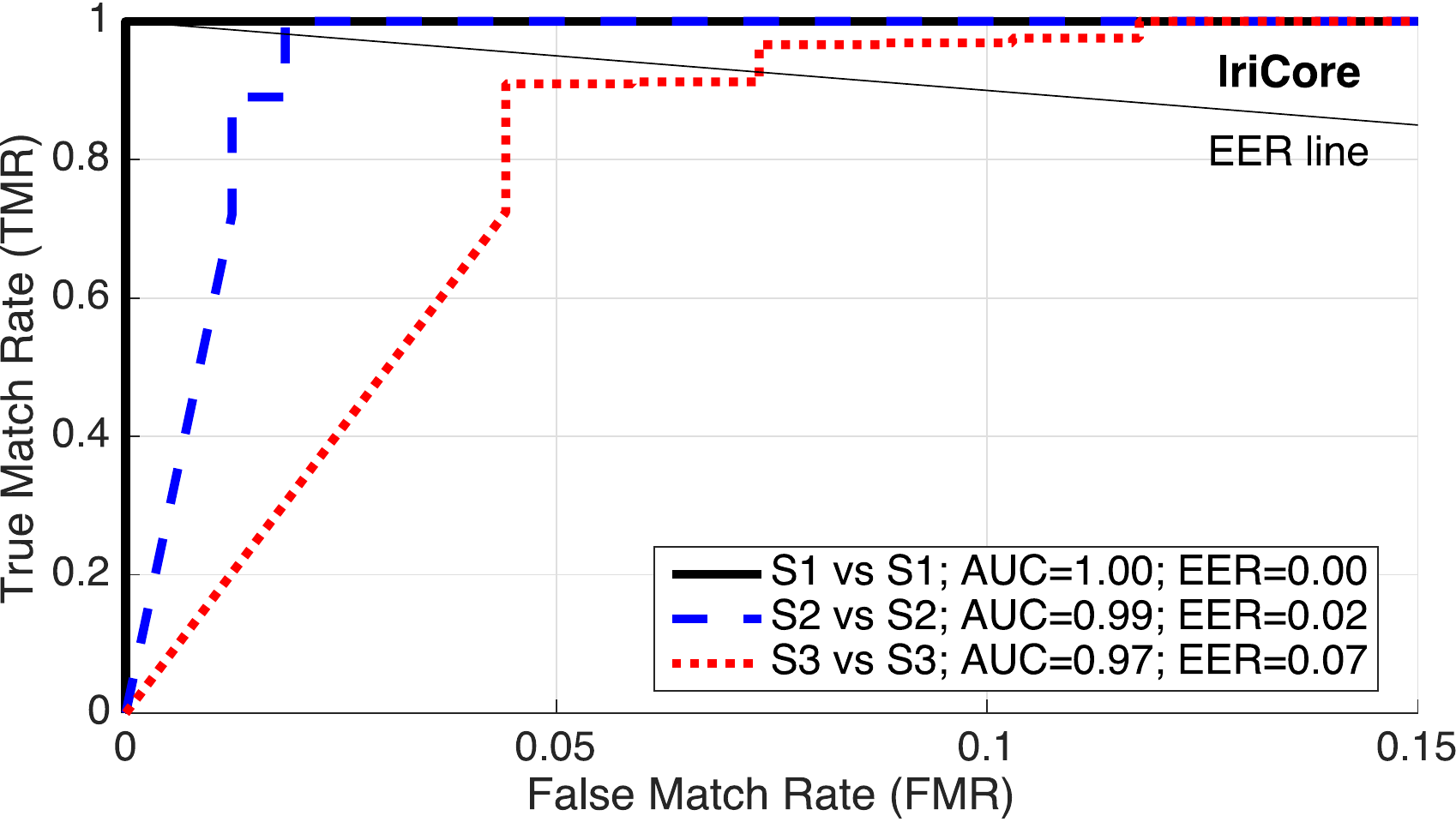}\hfill
	\includegraphics[width=0.46\textwidth]{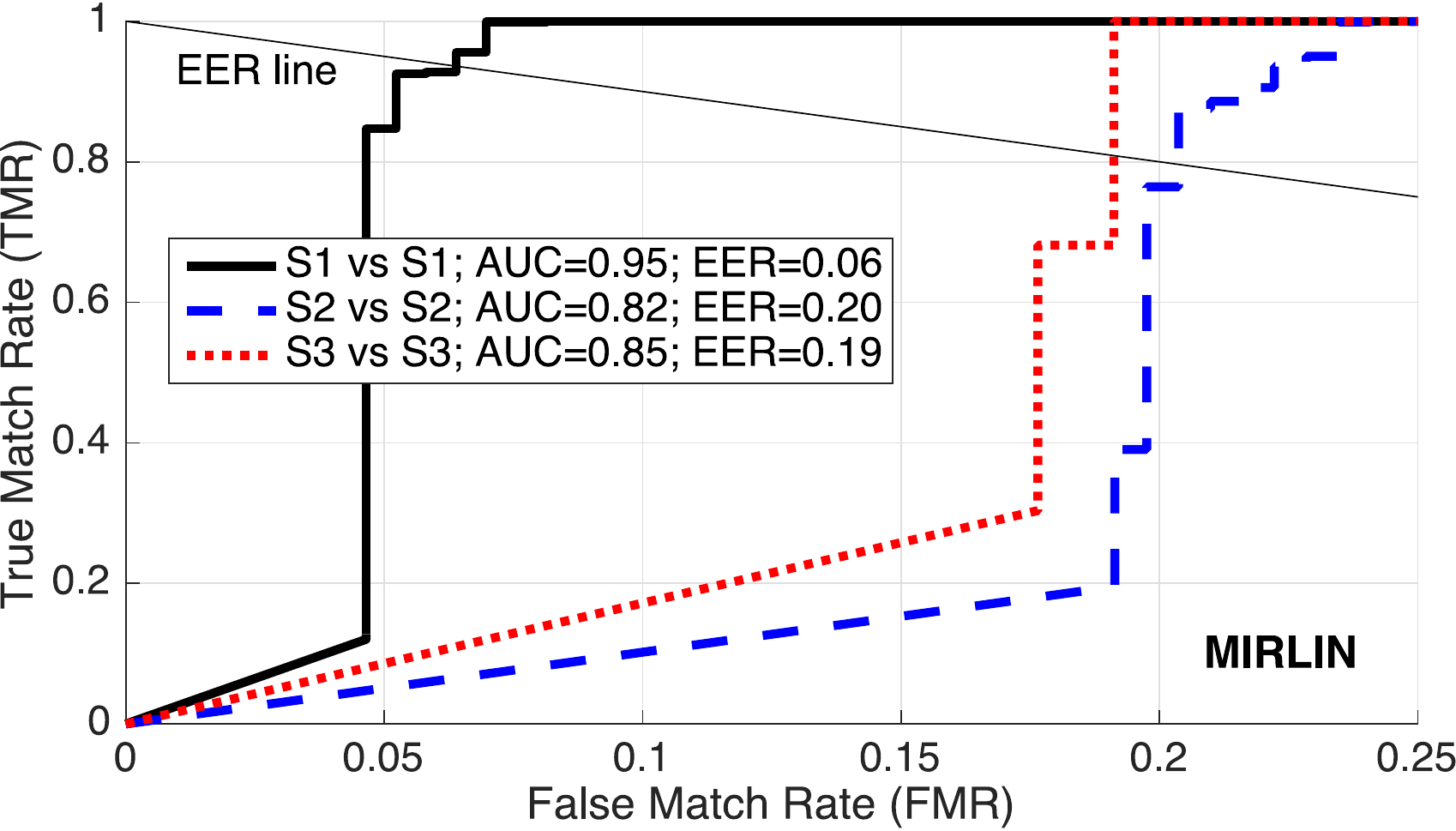}\\
	\includegraphics[width=0.46\textwidth]{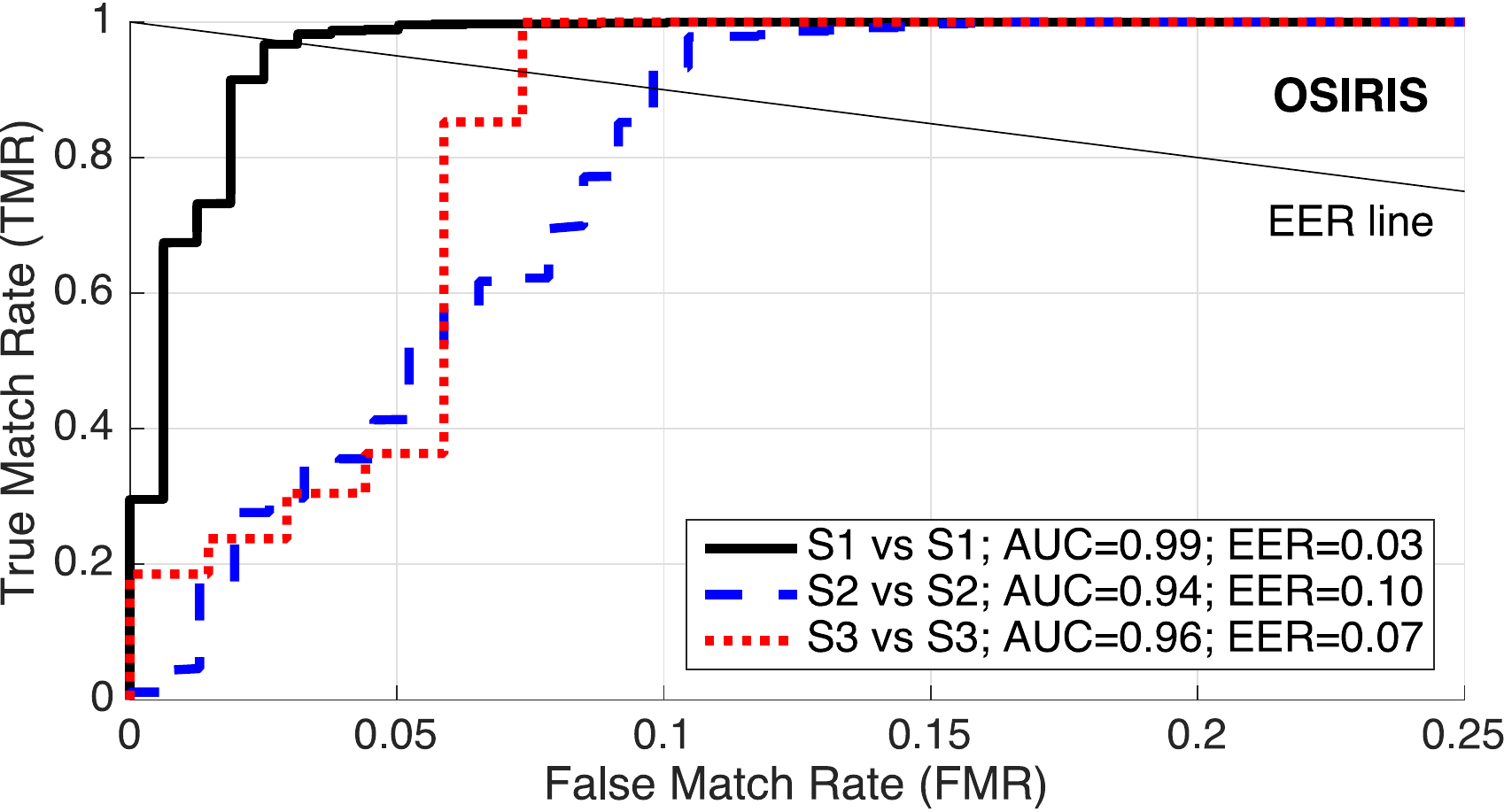}\hfill
	\includegraphics[width=0.46\textwidth]{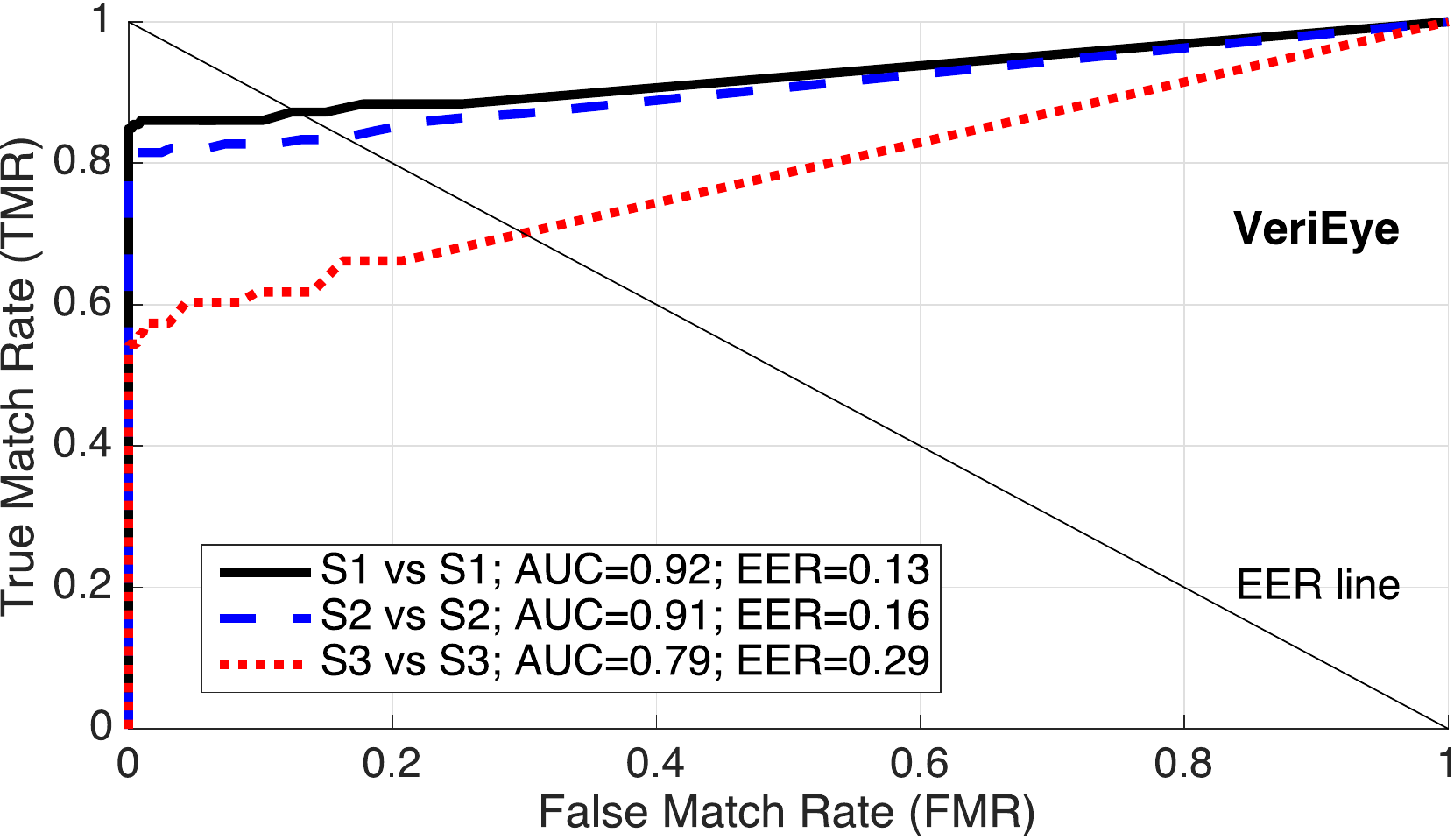}
	\caption{ROC curves for scores obtained when matching same-session samples (session 1: solid black line, session 2: dashed blue line, session 3: red dotted line). Area under curve (AUC) and equal error rate (EER) are also shown.}
	\label{fig:sameSessions}
\end{figure*}

Short-term analysis is performed with samples from the first three sessions, organized up to 60 hours after death. Due to sufficient number of samples, both genuine and impostor comparisons were calculated and presented in a form of ROC curves. Also, both intra- and inter-session analyses were performed. With difficulty to obtain ante-mortem and then post-mortem iris scans from the same individuals (mostly due to ethical concerns), the approximation used in this work is based on assumption that iris scans obtained shortly after death would not differ much from those obtained ante-mortem. Thus, session 1 images serve as enrollment samples in all analyzes.


%
%

Fig. \ref{fig:sameSessions} illustrates {\bf intra-session analysis}. All four matchers present worse performance in sessions 2 and 3 than in session 1. However, IriCore surprisingly achieves perfect recognition for session 1 samples with EER no larger than 7\% for irises photographed 60 hours after death. In general, the worst EER for samples acquired in such long period after death is 29\%, which is better than 50\% expected for random classifier, indicating a possibility to recognize large subset of cadaver samples. To check statistical significance of differences in genuine and impostor scores, one-tailed and two-sample Kolmogorov-Smirnov test at $\alpha=0.05$ was used. Tab. \ref{magicTable} details the results and alternative hypotheses (different across matchers due to different meaning of the comparison scores). All (except for OSIRIS genuine comparisons) intra-session score distributions revealed statistically significant differences between sessions (green color in `Genuine comparison scores' section denotes alternative hypotheses acceptance). Impostor scores revealed lower changes, since 50\% of null hypotheses on equality of distributions were rejected. However, rejection of null hypotheses for impostor scores indicates little risk of increasing false matches among post-mortem irises.

%
%
%
%
%
%
%

\begin{figure*}[!htb]
	\centering
	\includegraphics[width=0.46\textwidth]{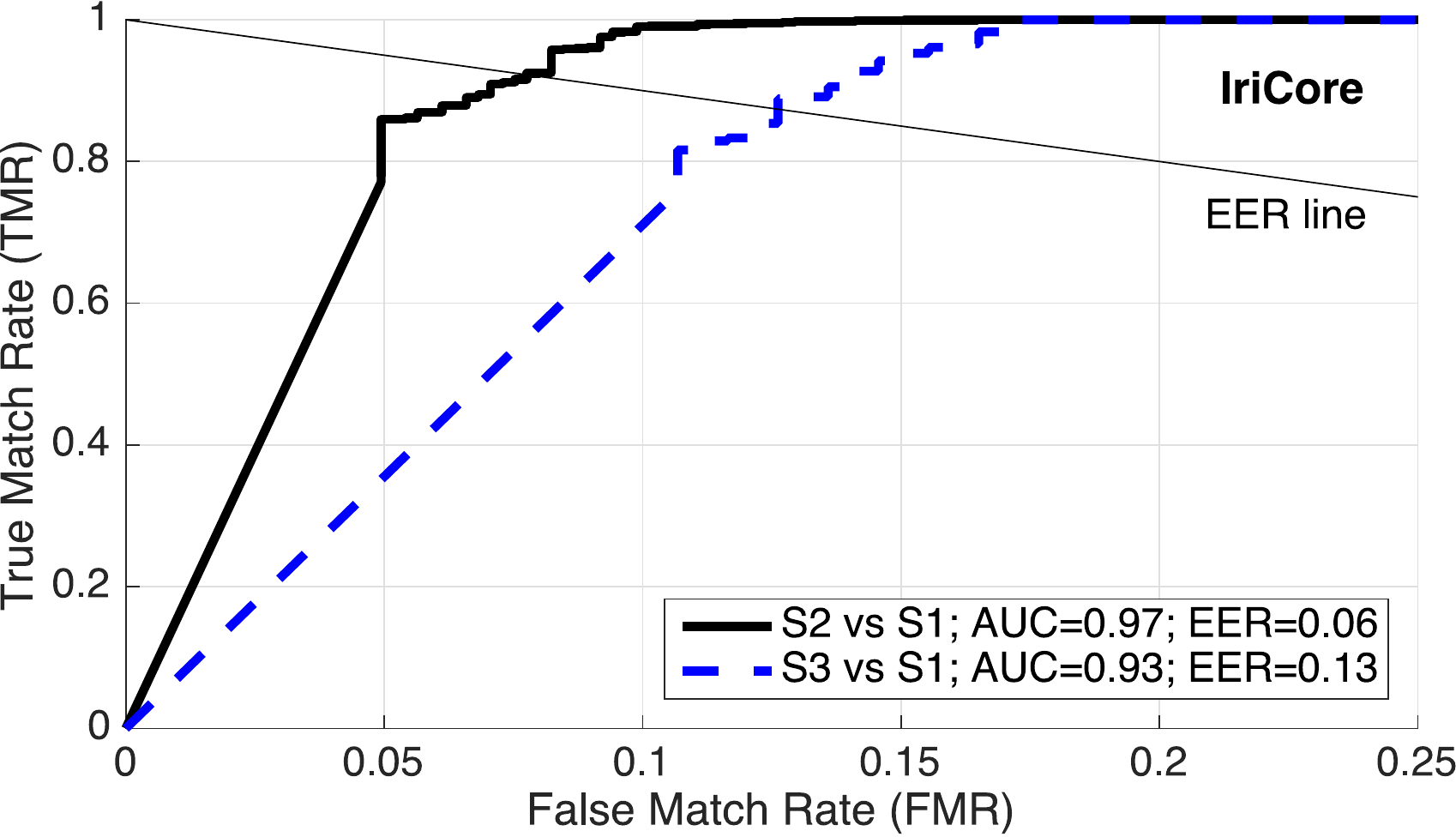}\hfill
	\includegraphics[width=0.46\textwidth]{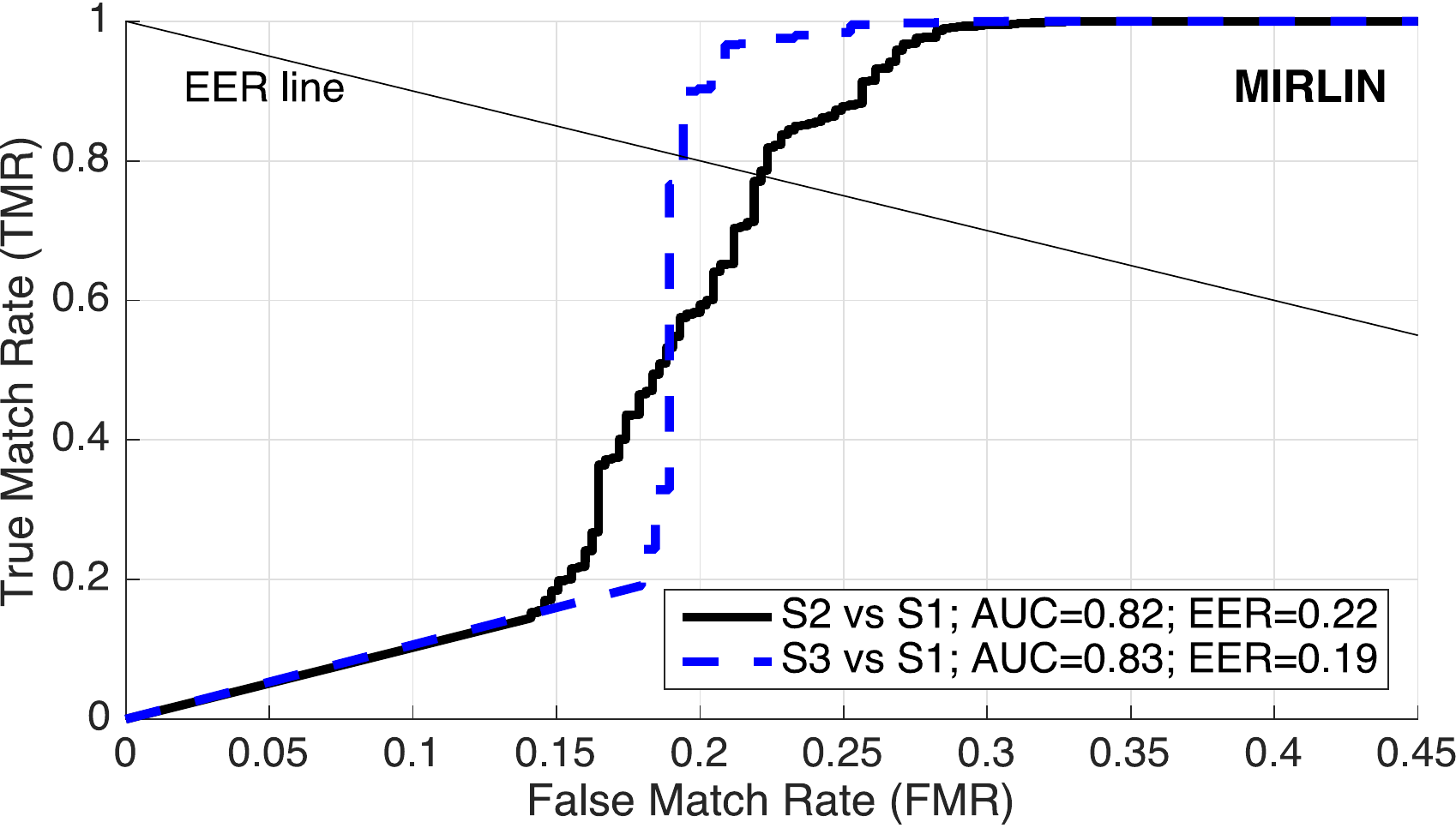}\\
	\includegraphics[width=0.46\textwidth]{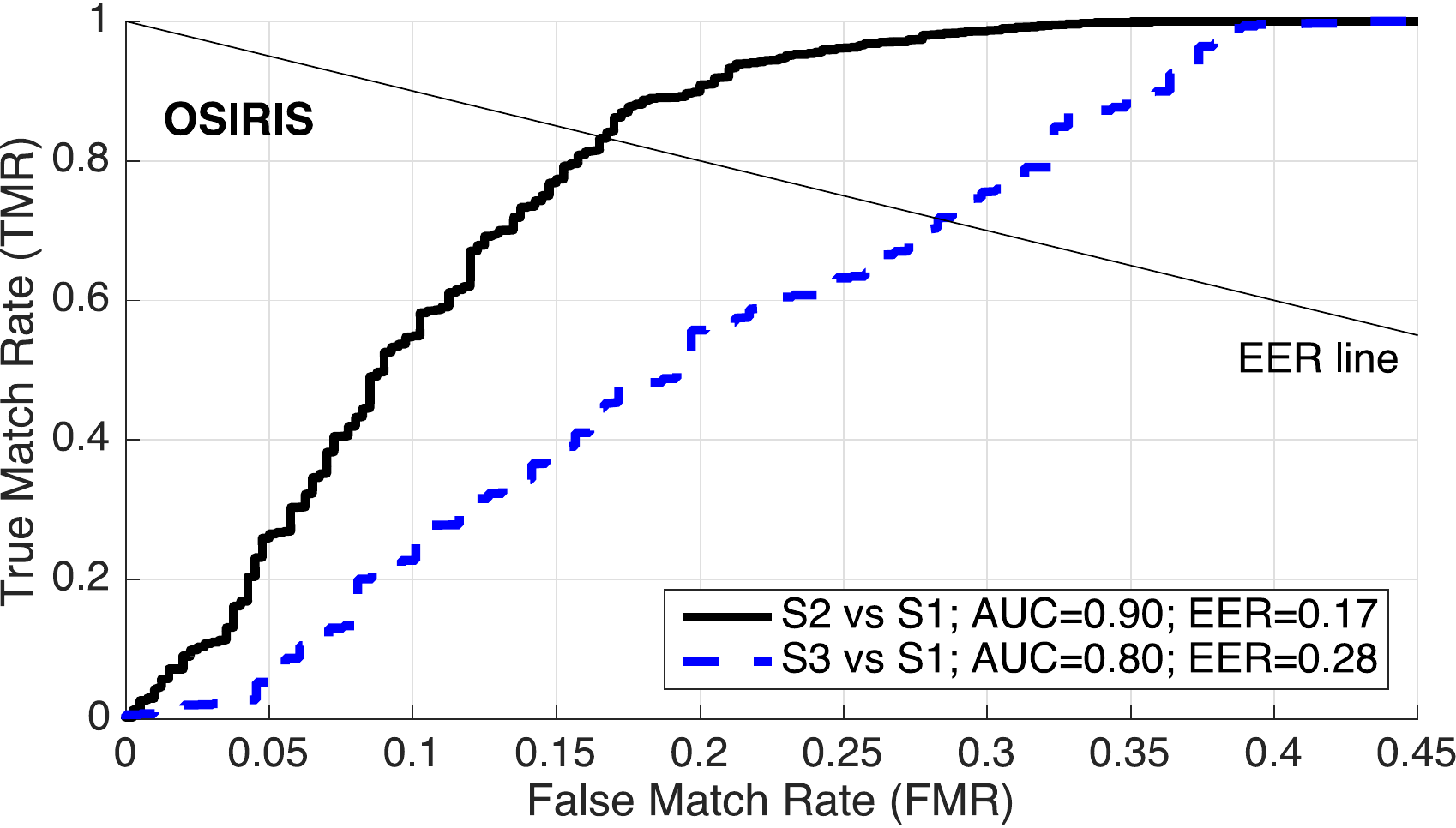}\hfill
	\includegraphics[width=0.46\textwidth]{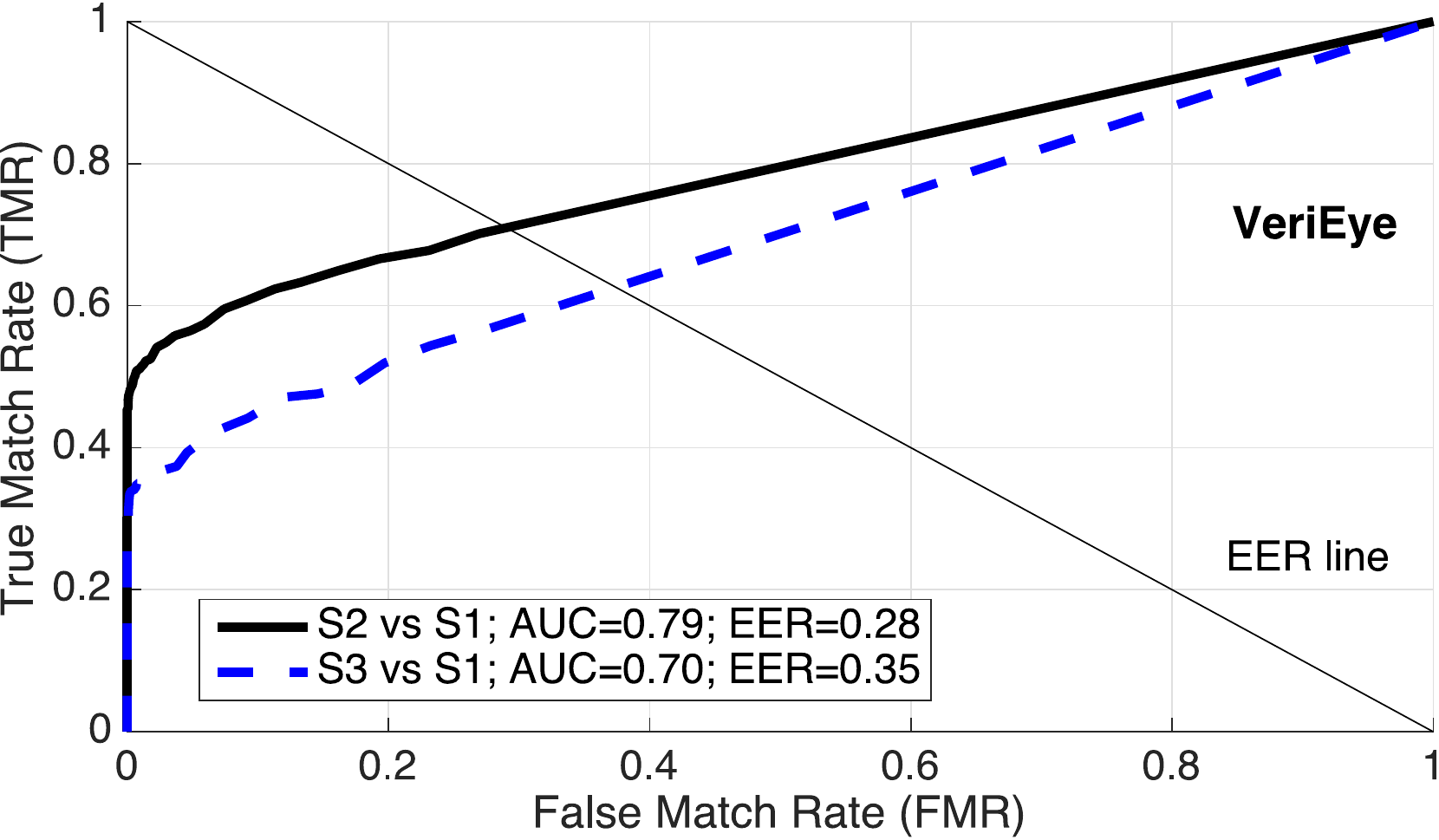}
	\caption{ROC curves for scores obtained when matching session 1 with session 2 images (solid black line), and session 1 with session 3 images (dashed blue line). Area under curve (AUC) and equal error rate (EER) are also shown.}
\label{fig:differentSessions}
\end{figure*}

Fig. \ref{fig:differentSessions} presents {\bf inter-session} analysis, where session 1 samples are compared with session 2, and also with session 3 samples (thus two ROC curves in each graph). All (except for MIRLIN) methods present worse performance for session 3 samples. However, IriCore's results suggest that comparison of samples taken up to 60 hours after demise and compared with those acquired 5--7 hours after death should lead to surprisingly low EER=13\%. Again, statistical significance of the genuine and impostor scores was analyzed, Tab. \ref{magicTable}. For all matchers we get statistically significant differences in genuine score distributions, and no statistically significant differences in impostor scores.

\begin{table*}[!htb]
\renewcommand{\arraystretch}{1.1}

\caption{Summary of statistical testing (Kolmogorov-Smirnov one-tailed and two-sample test at $\alpha=0.05$) of differences among distributions of genuine and impostor scores separately. $g$ and $i$ denote cumulative distribution functions (CDF) for genuine and impostor scores, respectively. The null hypotheses H0 in each test state that compared samples come from the same distribution. Alternative hypotheses H1 are defined in rows labeled `H1', and they are marked in red when there was no reason to reject H0 ($p$-value $>\alpha$) or in green when the null hypothesis was rejected and the corresponding alternative hypothesis was selected ($p$-value $\leq\alpha$). $g_{mn} > g_{lk}$ means that CDF of genuine scores between sessions $m$ and $n$ is shifted to \emph{lower} comparison scores when compared to the CDF of scores between sessions $l$ and $k$, \ie the plot of $g_{mn}$ is \emph{higher} than the plot of $g_{lk}$ (analogously for impostor scores $i$). Note opposite definitions of alternative hypotheses for VeriEye, since this method calculates similarity scores, while the remaining methods deliver dissimilarity scores.}

\label{magicTable}
\label{table:statTests}
\centering\scriptsize
\begin{tabular}[t]{cc|c|c|c|c|c||c|c|c|c|c|}
\cline{3-12}
& & \multicolumn{5}{c||}{\bf Genuine comparison scores} & \multicolumn{5}{c|}{\bf Impostor comparison scores} \\
\cline{3-12}
& & S1 vs S1 & S2 vs S2 & S3 vs S3 & S2 vs S1 & S3 vs S1  & S1 vs S1 & S2 vs S2 & S3 vs S3 & S2 vs S1 & S3 vs S1  \\\hline\hline
\multicolumn{1}{|c}{\multirow{3}{*}{IriCore}} & \multicolumn{1}{|c|}{\bf mean} & 0.642 & 0.703 & 0.757 & 0.944 & 1.065 & 1.863 & 1.811 & 1.812 & 1.848 & 1.839 \\\cline{2-12}
\multicolumn{1}{|c}{} & \multicolumn{1}{|c|}{\bf H1} &\cellcolor{gray!20}& \cellcolor{green!20} $g_{11} > g_{22}$  & \cellcolor{green!20} $g_{11} > g_{33}$  & \cellcolor{gray!20} & \cellcolor{green!20} $g_{21} > g_{31}$ & \cellcolor{gray!20} & \cellcolor{green!20} $i_{11} < i_{22}$  & \cellcolor{green!20} $i_{11} < i_{33}$  & \cellcolor{gray!20} & \cellcolor{red!20} $i_{21} < i_{31}$  \\\cline{2-12}
\multicolumn{1}{|c}{} & \multicolumn{1}{|c|}{\bf $p$-value} &\cellcolor{gray!20} & 0.018 & 0.002 & \cellcolor{gray!20} & \texttildelow 0 & \cellcolor{gray!20} & \texttildelow 0 & \texttildelow 0 &\cellcolor{gray!20} & 0.281\\\hline

\multicolumn{1}{|c}{\multirow{3}{*}{MIRLIN}} & \multicolumn{1}{|c|}{\bf mean} & 0.086 & 0.229 & 0.203 & 0.264 & 0.290 & 0.465 & 0.510 & 0.577 & 0.471 & 0.505\\\cline{2-12}
\multicolumn{1}{|c}{} & \multicolumn{1}{|c|}{\bf H1} & \cellcolor{gray!20} & \cellcolor{green!20} $g_{11} > g_{22}$  \cellcolor{green!20} & \cellcolor{green!20} $g_{11} > g_{33}$  & \cellcolor{gray!20} & \cellcolor{green!20} $g_{21} > g_{31}$ & \cellcolor{gray!20} & \cellcolor{red!20} $i_{11} < i_{22}$  & \cellcolor{red!20} $i_{11} < i_{33}$  & \cellcolor{gray!20} & \cellcolor{red!20} $i_{21} < i_{31}$  \\\cline{2-12}
\multicolumn{1}{|c}{} & \multicolumn{1}{|c|}{\bf $p$-value} & \cellcolor{gray!20} & 0.004 & 0.016 & \cellcolor{gray!20}  & \texttildelow 0 & \cellcolor{gray!20} & 0.749 & 0.983 & \cellcolor{gray!20}  & 0.154\\\hline

\multicolumn{1}{|c}{\multirow{3}{*}{OSIRIS}} & \multicolumn{1}{|c|}{\bf mean} & 0.225 & 0.247 & 0.247 & 0.336 & 0.380 & 0.461 & 0.458 & 0.461 & 0.460 & 0.460\\\cline{2-12}
\multicolumn{1}{|c}{} & \multicolumn{1}{|c|}{\bf H1} & \cellcolor{gray!20} & \cellcolor{red!20} $g_{11} > g_{22}$  & \cellcolor{green!20} $g_{11} > g_{33}$  & \cellcolor{gray!20} & \cellcolor{green!20} $g_{21} > g_{31}$ & \cellcolor{gray!20} & \cellcolor{green!20} $i_{11} < i_{22}$  & \cellcolor{red!20} $i_{11} < i_{33}$  & \cellcolor{gray!20} & \cellcolor{red!20} $i_{21} < i_{31}$  \\\cline{2-12}
\multicolumn{1}{|c}{} & \multicolumn{1}{|c|}{\bf $p$-value} & \cellcolor{gray!20} & 0.140 & 0.035 & \cellcolor{gray!20}  & \texttildelow 0 & \cellcolor{gray!20} & \texttildelow 0  & 0.052 & \cellcolor{gray!20}  & 0.193 \\\hline

\multicolumn{1}{|c}{\multirow{3}{*}{VeriEye}} & \multicolumn{1}{|c|}{\bf mean} & 282 & 234 & 117 & 82.7 & 33.9 & 1.44 & 1.85 & 1.76 & 1.58 & 1.64 \\\cline{2-12}
\multicolumn{1}{|c}{} & \multicolumn{1}{|c|}{\bf H1} & \cellcolor{gray!20} & \cellcolor{green!20} $g_{11} < g_{22}$  \cellcolor{green!20} & \cellcolor{green!20} $g_{11} < g_{33}$  & \cellcolor{gray!20} & \cellcolor{green!20} $g_{21} < g_{31}$ & \cellcolor{gray!20} & \cellcolor{green!20} $i_{11} > i_{22}$  & \cellcolor{red!20} $i_{11} > i_{33}$  & \cellcolor{gray!20} & \cellcolor{red!20} $i_{21} > i_{31}$  \\\cline{2-12}
\multicolumn{1}{|c}{} & \multicolumn{1}{|c|}{\bf $p$-value} & \cellcolor{gray!20} & 0.001 & \texttildelow 0 & \cellcolor{gray!20}  & \texttildelow 0 &\cellcolor{gray!20}  & \texttildelow 0 & 0.088 & \cellcolor{gray!20}  & 0.616 \\\hline

\end{tabular}
\end{table*}

\subsection{Long-term analysis}

\begin{figure*}[!htb]
	\centering
	\includegraphics[width=0.47\textwidth]{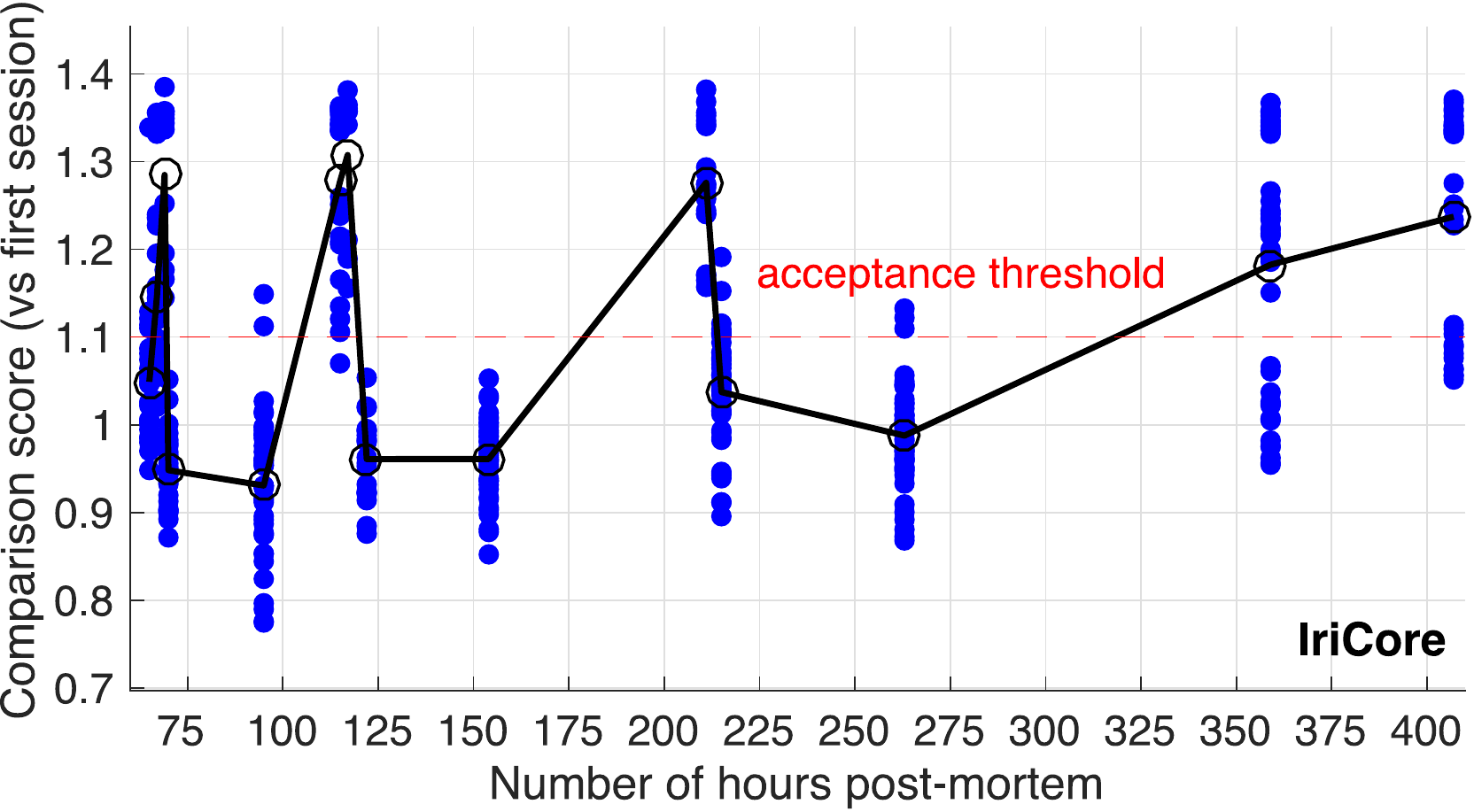}\hfill
	\includegraphics[width=0.47\textwidth]{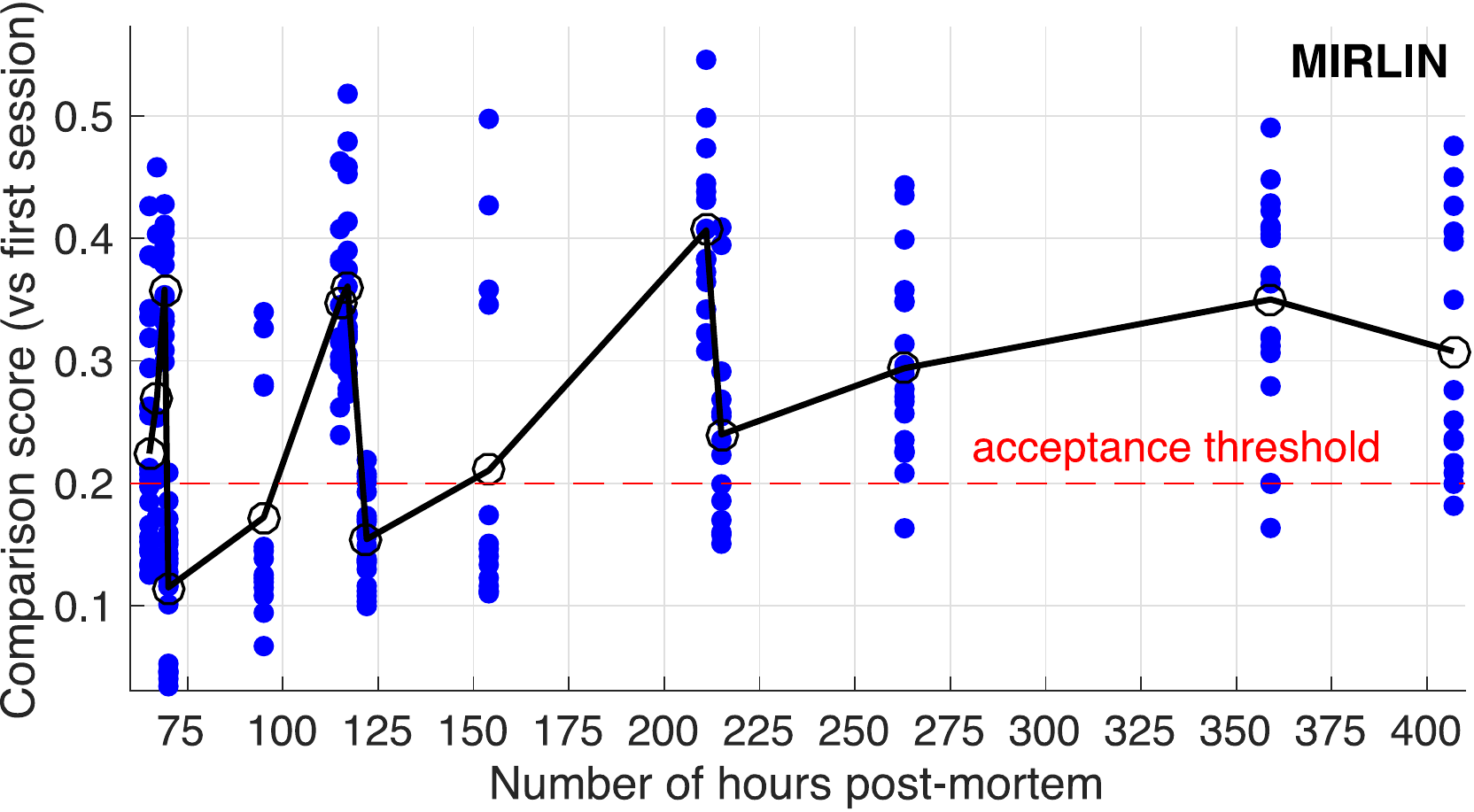}\\
	\includegraphics[width=0.47\textwidth]{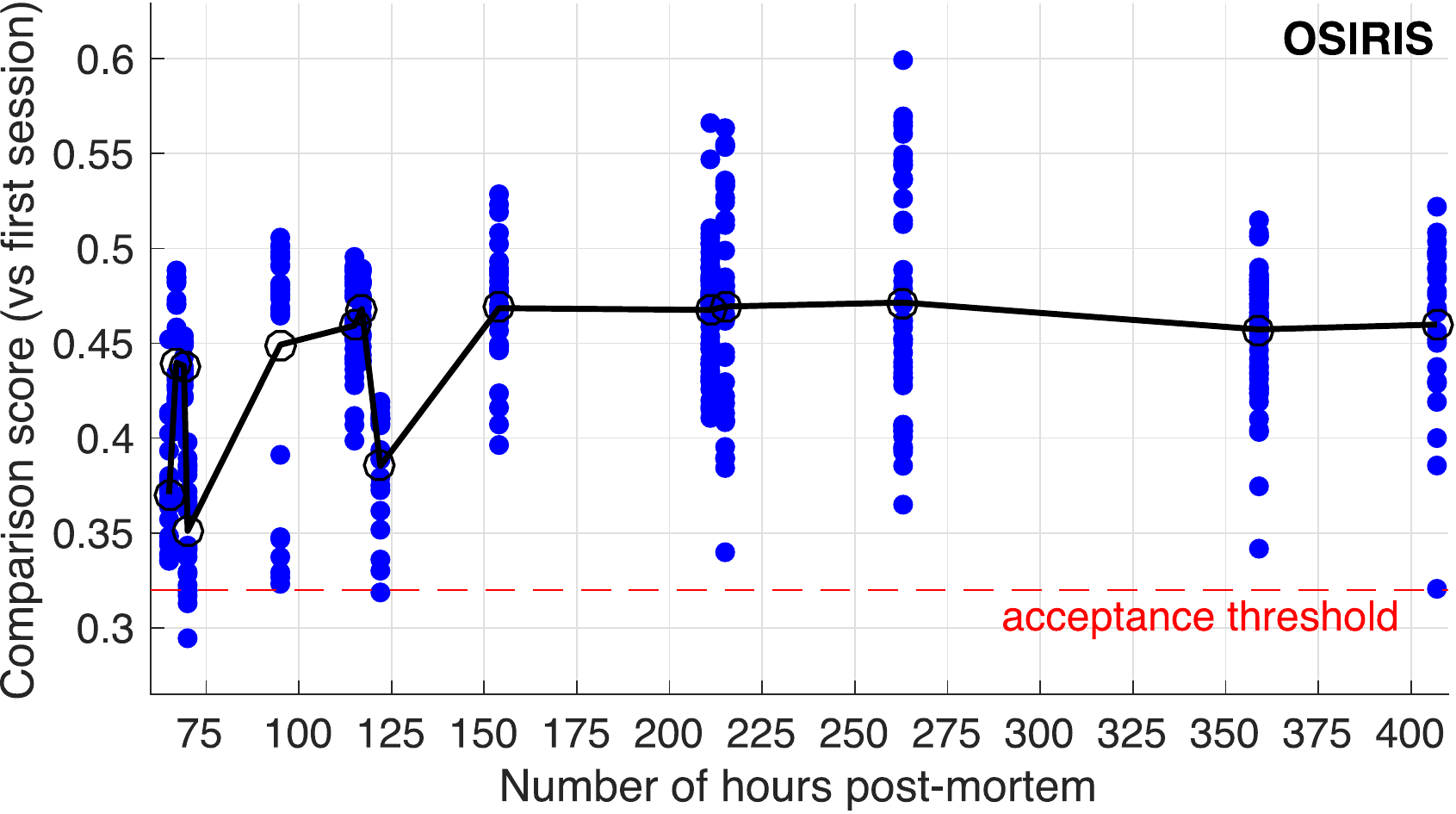}\hfill
	\includegraphics[width=0.47\textwidth]{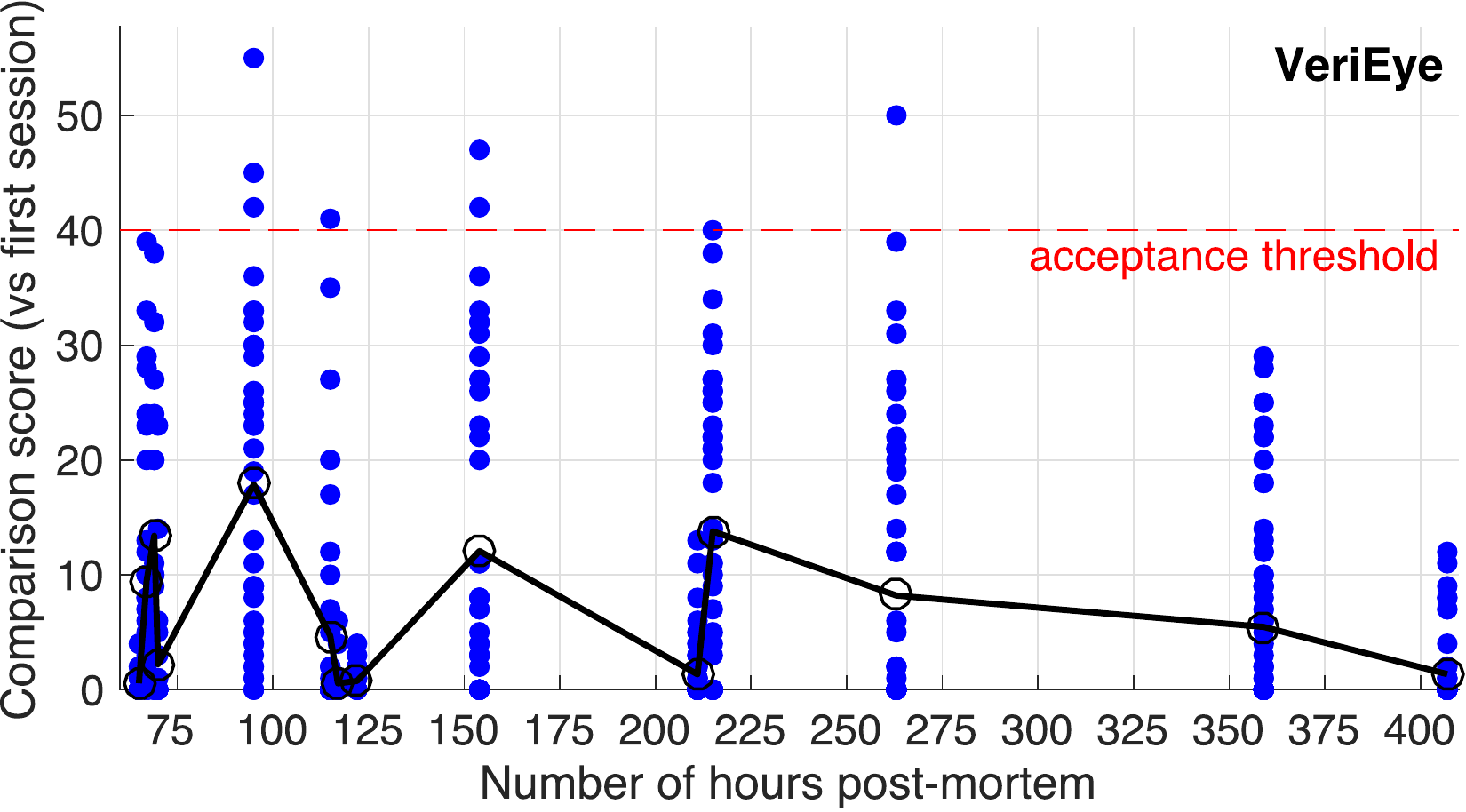}
	\caption{Long-term analysis. Blue dots represent comparison scores between samples acquired in the first session (\ie, 5-7 hours after decease) and samples acquired at least 60 hours after decease (as denoted on horizontal axis). Black circles are means for a given time distance between acquisitions. Scores {\bf below} the acceptance threshold denote correct matches for the \textbf{IriCore}, \textbf{MIRLIN}, and \textbf{OSIRIS} methods, while scores \textbf{above} the threshold denote correct matches for the \textbf{VeriEye} method.}
\label{fig:longTerm}
\end{figure*}

Samples acquired more than 60 hours after death are sparsely distributed in time and across the subjects, and thus are not adequate for an analysis such as that done for short-term samples. The basic question, however, is how long after death the iris can still offer individual features, enough to offer a correct match, and this long-term analysis of genuine scores provides an important answer. Fig. \ref{fig:longTerm} presents all genuine scores calculated between session 1 images (5--7 hours after death) and all the samples acquired more than 60 hours after demise. Default acceptance thresholds are shown in red to estimate how many samples (and up to what time horizon) can still be correctly recognized. For IriCore and MIRLIN we occasionally get correct matches for samples acquired 407 hours after death. VeriEye occasionally recognizes samples acquired up to 260 hours after demise and OSIRIS gives correct matches after up to 124 hours, at the assumed acceptance thresholds. Results show that iris recognition reliability is severely deteriorated after such periods, yet surprisingly still possible.

%
%
%

\section{Conclusions}

This paper brings two important deliverables: a) a comprehensive study of human port-mortem iris recognition, carried out for four different matchers and encompassing samples acquired up to 17 days, and b) a unique database of NIR and visible light images to make further research in this area possible by others. The main conclusion of this study is that post-mortem iris recognition is possible, and the equal error rate may be as low as 13\% (IriCore's result for inter-session comparisons) if samples acquired up to 60 hours after death are compared to those captured 5--7 hours post-mortem. The same matcher also gives a perfect recognition accuracy for the latter set of samples. One may still expect occasional correct matches of samples collected almost 17 days after demise. We hope that this paper will start a scientific debate on post-mortem automatic human iris recognition and, together with the offered database, inspire further research in this important area.


{\small
\bibliographystyle{ieee}
\bibliography{refs}
}

\end{document}